\pdfoutput=1

\documentclass[11pt]{article}

\usepackage{ACL2023}

\usepackage{times}
\usepackage{latexsym}
\usepackage{amsmath}
\usepackage{amssymb}

\usepackage{mathrsfs}

\DeclareFontFamily{OT1}{pzc}{}
\DeclareFontShape{OT1}{pzc}{m}{it}{<-> s * [1.10] pzcmi7t}{}
\DeclareMathAlphabet{\mathpzc}{OT1}{pzc}{m}{it}

\usepackage{pifont}
\newcommand{\cmark}{\ding{51}}%
\newcommand{\xmark}{\ding{55}}%

\usepackage[T1]{fontenc}

\usepackage[utf8]{inputenc}

\usepackage{microtype}

\usepackage{inconsolata}

\usepackage{color, colortbl}
\usepackage{booktabs}
\usepackage{multirow}
\usepackage{makecell}

\usepackage{hyperref}
\usepackage{url}

\usepackage{graphicx}
\usepackage{booktabs} 

\usepackage{graphics}

\NewDocumentCommand{\heng}
{ mO{} }{\textcolor{red}{\textsuperscript{\textit{Heng}}\textsf{\textbf{\small[#1]}}}}
\NewDocumentCommand{\mbc}
{ mO{} }{\textcolor{purple}{\textsuperscript{\textit{Mohit}}\textsf{\textbf{\small[#1]}}}}
\NewDocumentCommand{\manling}
{ mO{} }{\textcolor{blue}{\textsuperscript{\textit{Manling}}\textsf{\textbf{\small[#1]}}}}
\NewDocumentCommand{\zoey}
{ mO{} }{\textcolor{orange}{\textsuperscript{\textit{Zoey}}\textsf{\textbf{\small[#1]}}}}
\NewDocumentCommand{\yu}
{ mO{} }{\textcolor{green}{\textsuperscript{\textit{Yu}}\textsf{\textbf{\small[#1]}}}}
\NewDocumentCommand{\xd}
{ mO{} }{\textcolor{blue}{\textsuperscript{\textit{Xudong}}\textsf{\textbf{\small[#1]}}}}
\NewDocumentCommand{\zhenhailong}
{ mO{} }{\textcolor{teal}{\textsuperscript{\textit{Zhenhailong}}\textsf{\textbf{\small[#1]}}}}

\usepackage{titlesec}

\title{Non-Sequential Graph Script Induction via Multimedia Grounding}

\author{Yu Zhou$^1$, Sha Li$^2$, Manling Li$^2$, Xudong Lin$^3$, Shih-Fu Chang$^3$, Mohit Bansal$^4$, Heng Ji$^2$\\
$^1$ University of California, Los Angeles
$^2$ University of Illinois Urbana-Champaign \\
$^3$ Columbia University
$^4$ University of North Carolina at Chapel Hill\\
\texttt{yu.zhou@ucla.edu}, \texttt{\{shal2, manling2, hengji\}@illinois.edu} \\ 
\texttt{mbansal@cs.unc.edu}, \texttt{\{xudong.lin, shih.fu.chang\}@columbia.edu} 
}

\begin{document}
\maketitle

\begin{abstract}
Online resources such as wikiHow compile a wide range of scripts for performing everyday tasks, which can assist models in learning to reason about procedures.~\footnote{Our data and code are publicly available for research purposes at \url{https://github.com/bryanzhou008/Multimodal-Graph-Script-Learning/}} 
However, the scripts are always presented in a linear manner, which does not reflect the flexibility displayed by people executing tasks in real life. 
For example, in the CrossTask Dataset, 64.5\% of consecutive step pairs are also observed in the reverse order, suggesting their ordering is not fixed. In addition, each step has an average of 2.56 frequent\footnote{Occurred in more than 10 videos.} next steps, demonstrating "branching".  
In this paper, we propose a new challenging task of non-sequential graph script induction, aiming to capture \textit{optional} and \textit{interchangeable} steps in procedural planning. To automate the induction of such graph scripts for given tasks, we propose to take advantage of loosely aligned videos of people performing the tasks. In particular, we design a multimodal framework to ground procedural videos to wikiHow textual steps and thus transform each video into an observed step path on the latent ground truth graph script.
This key transformation enables us to train a script knowledge model capable of both generating explicit graph scripts for learnt tasks and predicting future steps given a partial step sequence. Our best model outperforms the strongest pure text/vision baselines by 17.52\% absolute gains on F\textsubscript{1}@3 for next step prediction and 13.8\% absolute gains on Acc@1 for partial sequence completion. Human evaluation shows our model outperforming the wikiHow linear baseline by 48.76\% absolute gains in capturing sequential and non-sequential step relations.

\end{abstract}



\section{Introduction} \label{intro_section}
A script consists of typical actions that are performed to complete a given task. Online resources such as wikiHow\footnote{\url{www.wikiHow.com}} provide a wide variety of community-edited scripts for everyday tasks (Fig.\ref{fig:schema}). 
Such a large library of linear scripts can serve as a starting point for learning goal-step knowledge~\cite{zhang-etal-2020-reasoning,yang-etal-2021-visual}. 
However, as the saying goes, ``all roads lead to Rome''. There is usually more than one way to achieve any given goal. Practically speaking, users should be presented with multiple alternative step sequences so that they can pick the most suitable route according to their unique situations and preferences. 
Robots and virtual assistants also stand to gain the crucial abilities of global planning optimization and on-the-spot improvisation from alternative step paths.

\definecolor{babypink}{rgb}{0.96, 0.76, 0.76}
\definecolor{lavenderblush}{rgb}{1.0, 0.94, 0.96}
\definecolor{mistyrose}{rgb}{1.0, 0.89, 0.88}
\definecolor{lightthulianpink}{rgb}{0.9, 0.56, 0.67}
\definecolor{mauvelous}{rgb}{0.94, 0.6, 0.67}
\definecolor{floralwhite}{rgb}{1.0, 0.98, 0.94}
\definecolor{moccasin}{rgb}{0.98, 0.92, 0.84}
\definecolor{lightgoldenrodyellow}{rgb}{0.98, 0.98, 0.82}
\definecolor{harvestgold}{rgb}{0.85, 0.57, 0.0}
\definecolor{gamboge}{rgb}{0.89, 0.61, 0.06}
\definecolor{goldenbrown}{rgb}{0.6, 0.4, 0.08}
\begin{figure}[t]
\begin{center}
\includegraphics[width=75mm,scale=0.5]{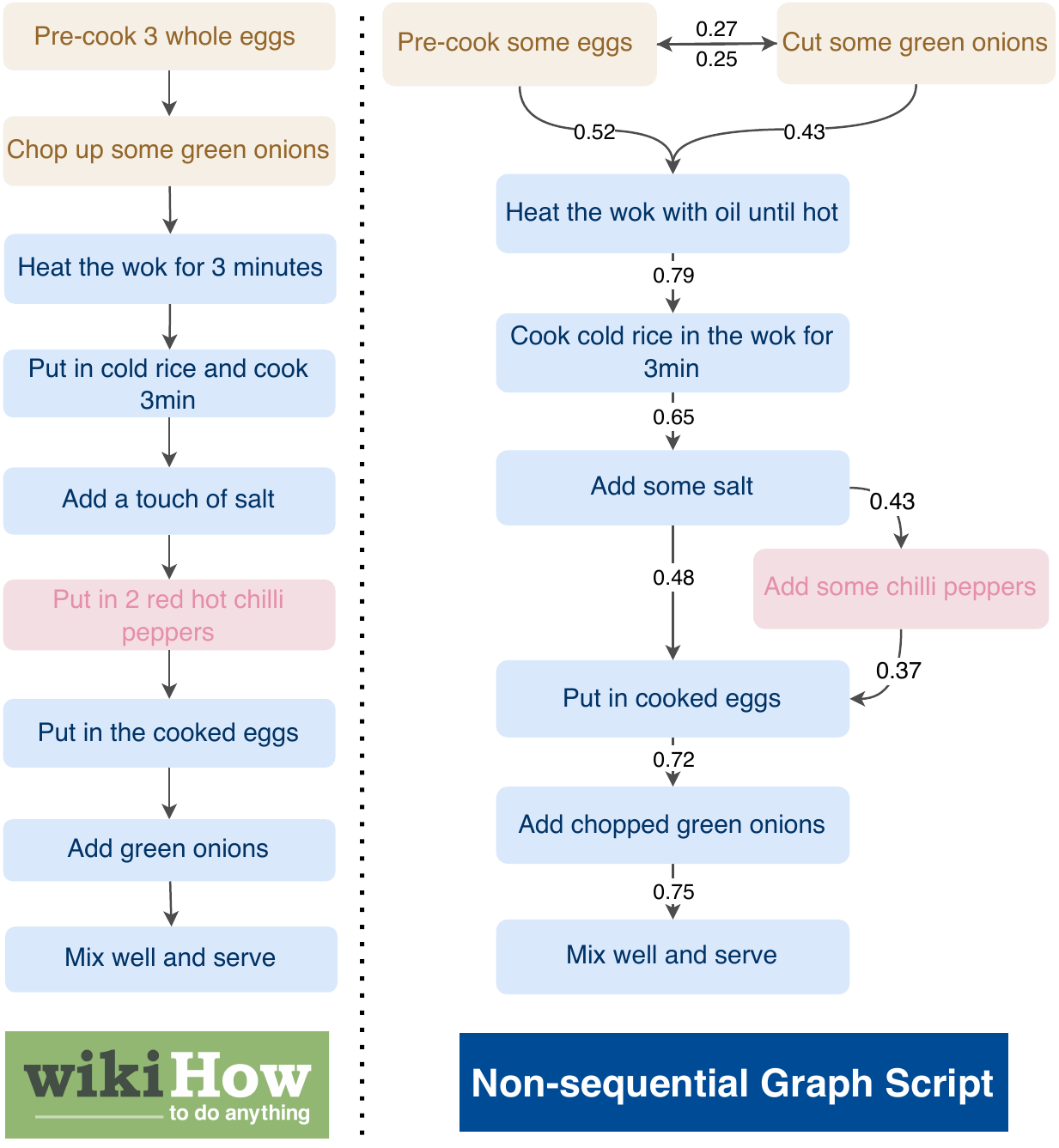}
\end{center}
\caption{Example of wikiHow linear script of the procedural task \textit{make egg fried rice} compared to an ideal example of our non-sequential graph script consisting of \colorbox{lavenderblush}{\color{lightthulianpink} optional} and \colorbox{moccasin}{\color{goldenbrown} interchangeable}
steps. 
}
\label{fig:schema}
\end{figure}

In particular, we observe that two types of steps are overlooked by linear scripts: \textit{optional steps} and \textit{interchangeable steps}. 
Optional steps such as \texttt{Add some chili peppers} can be skipped based on the users' preference or item availability. Interchangeable steps such as \texttt{Pre-cook some eggs} and \texttt{Cut some green onions} can be performed in either order without affecting the overall task completion.
After accounting for these two step types, the original linear script is converted into a `non-sequential graph script', as shown in Fig.\ref{fig:schema} (right).

\begin{figure*}[ht]
\begin{center}
\includegraphics[width=160mm,scale=1]{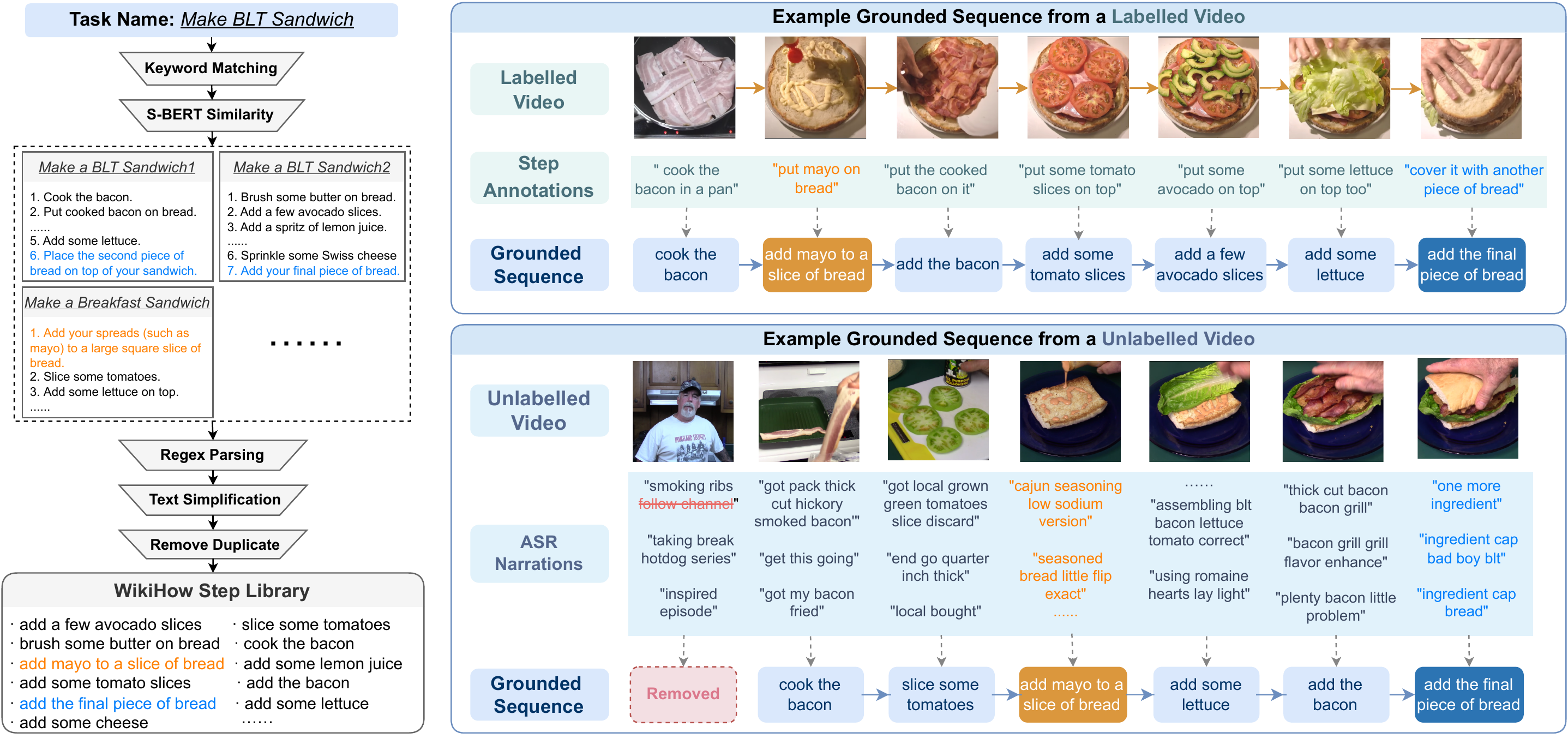}
\end{center}
\caption{Example of grounding procedural videos of the \texttt{Making BLT Sandwich} task to wikiHow steps. 
We  create a wikiHow step library through task-level matching and step pre-processing, and then ground video step annotations/asr-narrations to textual steps from the wikiHow step library. }
\label{fig:grounding}
\end{figure*}

Previous efforts like Proscript~\cite{sakaguchi-etal-2021-proscript-partially} obtained non-linear graph scripts via crowdsourcing, which is not scalable. In this work, we automate the process of transforming a linear text script into a non-linear graph script by grounding into visual observations (videos) of people executing the task. If we observe that people often skip a certain step, then it is natural to denote that step as optional. Similarly, if people tend to swap the ordering of a group of steps, these steps are likely interchangeable. Since wikiHow does not contain such emperical observations, we align wikiHow scripts with procedural video datasets such as
Crosstask~\cite{Zhukov2019CrossTaskWS} and Howto100M~\cite{Miech2019HowTo100MLA} (see Fig.\ref{fig:grounding}).

 To map a video to a sequence of wikiHow steps, we perform alignment on both task-level and step-level. 
 On the task level, we use a title matching algorithm based on Sentence-BERT similarity to select videos and wikiHow documents for the task. Then, we propose an effective pre-possessing strategy (simplification + deduplication) to create the wikiHow step libarary. 
 At the step level, we consider two situations based on whether the video has been segmented into steps. 
 When manual segmentation is provided, we directly map video annotations to the wikiHow step library. Otherwise, we first segment
 the video into clips based on ASR sentence groups (Fig.\ref{fig:grounding}), and then map them to wikiHow steps using a fault tolerant grounding strategy (\S\ref{grounding_section}) that is robust to inaccurate ASR sentence boundaries. 
 When grounding is complete, we obtain the set of observed step sequences for each task. 

 
 
 Next, to obtain the desired graph script from the observed step sequences, we use auto-regressive seq2seq models \cite{DBLP:journals/corr/SutskeverVL14} to learn the distribution of valid paths (step sequences) along the graph (\S\ref{Script learning}). 
 As opposed to directly training a graph generation model, our path generation learning format is better aligned with existing procedural video data
 and also takes advantage of pretrained seq2seq models to improve generalization across tasks. 
%
 Since the cross-entropy loss used for training auto-regressive models focuses on penalizing local ``one-step'' errors (the errors in predicting each single step), we further introduce a Path-level Constraint Loss to reduce global inconsistencies of the entire path. To generate hard negative contrastive-paths that fail to complete the task, we manipulate the video-grounded positive paths through \textit{global reordering}, \textit{shuffling}, and \textit{re-sampling} (\S\ref{contrastive_loss_section}).
 
%
 After training, our model is able to produce complete paths given input step libraries from various domains, including but not limited to: cooking, car maintenance, and handcrafting, etc. To automatically generate explicit
 graph scripts, we implement step-level constraint beam-decoding to sample multiple generated step sequences and record a step-adjacency matrix for constructing the final graph script.

 For downstream evaluation, we adapt the existing CrossTask dataset~\cite{Zhukov2019CrossTaskWS} to set up two new evaluation sub-tasks: \textit{Next Step Prediction} and \textit{Partial Sequence Completion}. Compared against top-performing test/video only baselines, our best model achieves 17.52\% absolute gains in overall F\textsubscript{1}@3 for next step prediction and 13.8\% absolute gains on Accuracy@1 for partial sequence completion. Moreover, we use MTurk to perform \textit{Human Evaluation} on the correctness and expressiveness of our auto-generated graph scripts. Results show our model can correctly capture optional, interchangeable and sequential step relationships with up to 82.69\% overall accuracy.




Key contributions of this paper include:
\begin{itemize}
    \item We introduce an automatic method for converting sequential text scripts into non-sequential graph scripts by aligning / grounding textual scripts to video datasets. 
    \item We propose a path generation model capable of 
    learning from video-grounded step sequences with Path-Level Constraint Loss.
    \item Experiments show our non-sequential path generation model to be more effective than existing text/vision baselines in next step prediction and partial sequence completion. 
    \item Human evaluation of generated graph scripts demonstrates our non-sequential graph scripts to be more accurate and expressive in capturing step-relationships.
\end{itemize}

\section{Task Formulation}

In this paper, we propose a new challenge of graph script induction for procedural tasks: Given a procedural task $\mathcal{T}$ represented by a task name, our goal is to induce a graph script for the task using the steps in the linear script. In particular, the graph script should capture the following relations between steps: (1) \textit{sequential} $\langle s_i \rightarrow s_j \rangle$ where two steps should be completed sequentially; (2) \textit{interchangeable} $\langle s_i \leftrightarrow s_j \rangle$ where two steps can be completed in either order or at the same time; (3) \textit{optional} $\langle s_i \rightarrow s_k,  s_i \rightarrow s_j \rightarrow s_k\rangle$ where a step can be optionally added between other steps. 

To achieve this goal, we assume that we have access to a large repository of textual scripts (wikiHow) and a set of videos that record people carrying out the tasks.\footnote{Or a large repository of videos from which we can find matching videos using retrieval.} The videos might have step-level annotations or accompanying narration which we can convert into text using ASR tools.


\section{Methodology}

To learn a graph script induction model, we first ground the video dataset to textual steps on both task-level and step-level (Fig.~\ref{fig:grounding}). After grounding, each video can be seen as a valid step sequence sampled from the ground truth graph script. 
Then, we use such grounded step sequences to train our graph script model and enhance model learning by introducing a Path-Level Constraint Loss over carefully designed contrastive step sequences.

\subsection{Video to Script Grounding} \label{grounding_section}

For each video, we first perform task-level alignment to find the top-$m$ most relevant wikiHow documents 
and then step-level alignment to ground the video to specific wikiHow steps.
We consider the following two cases based on whether the video dataset includes step-level annotation:

\paragraph*{Labelled Video Datasets:} Labelled video datasets like Crosstask~\cite{Zhukov2019CrossTaskWS} contain procedural videos grouped by human-annotated task names. In addition, the videos are labelled with temporal step segmentation and relatively accurate step annotations in the form of short imperative English sentences. The example video in Fig.\ref{fig:grounding} for task: \underline{\textit{"Make BLT Sandwich"}} is annotated with steps: "cook the bacon in a pan", "put mayo on bread", etc.

At the task level, we first use keyword matching to quickly find all relevant wikiHow documents whose title contains $\ge$ 85\% of keywords in the task name. For example in Fig.~\ref{fig:grounding}, the task name: \underline{\textit{"Make BLT Sandwich"}} is matched to wikiHow documents: \underline{\textit{"Make a BLT Sandwich1"}}, \underline{\textit{"Make a Breakfast Sandwich"}}, etc.
After we retrieve a list of relevant wikiHow documents, they are further ranked by cosine similarity between Sentence-BERT embeddings of document title and the task name. Finally, the steps of the top $m$ wikiHow documents are selected to form the initial wikiHow step candidate pool.

In step-level grounding, we first record Sentence-BERT Similarity scores between each video step annotation and all processed steps in the wikiHow step library. Then, we do greedy matching between video step annotations and wikiHow steps with priority given to higher scoring pairs. Here we keep video steps with best score $\geq k_1$\footnote{Hyperparameters in the grounding section are empirically selected based on qualitative evaluation over a small subset.}, while lower scoring video steps are considered ungroundable. When all videos have been grounded, unused steps from the wikiHow step library  are removed.



\paragraph*{Unlabelled Video Datasets:}
Although we achieve high grounding quality for annotated video datasets, step-level annotation is quite costly and often not available for a wide range of tasks that we are interested in. 
A more practical scenario is when we have a large repository of videos like Howto100M from which we can retrieve videos corresponding to the target task. 
%
Task-level alignment for Howto100M is different from that of annotated video datasets due to questionable video grouping. In Howto100M, videos for each task are selected purely based on Youtube search ranking. This ranking often prioritizes popular videos that have low correlation to the task at hand. 
To ensure high video-task correlation, we re-select Howto100M videos for each task based on BERT-Similarity between video title and the task name (only videos with similarity score $\geq k_2$ are selected). 

Step-level alignment also becomes much more challenging as we must rely on video ASR transcriptions without human step-level annotations. ASR narrations usually comprise of short partial sentence pieces without strict temporal step boundary labels (Fig.\ref{fig:grounding}). In addition, since Howto100M videos are collected from Youtube, some ASR narrations contain task-irrelevant information such as subscription requests (Fig.\ref{fig:grounding}). To address these challenges, we use a more fault tolerant grounding strategy shown in Fig.\ref{fig:grounding}: First, we remove all sentence pieces containing Youtube stop words including “subscribe”, “channel”, “sponsor”, etc.
Then, we expand each ASR sentence piece by concatenating it with surrounding pieces until the length of the resulting piece exceeds 10 words\footnote{This parameter is borrowed from~\cite{lin2022learning} which uses the same length threshold}. 
Finally, we ground each resulting ASR step to wikiHow steps with a higher match threshold $k_3$. 

\begin{figure*}[ht]
\begin{center}
\includegraphics[width=170mm,scale=1]{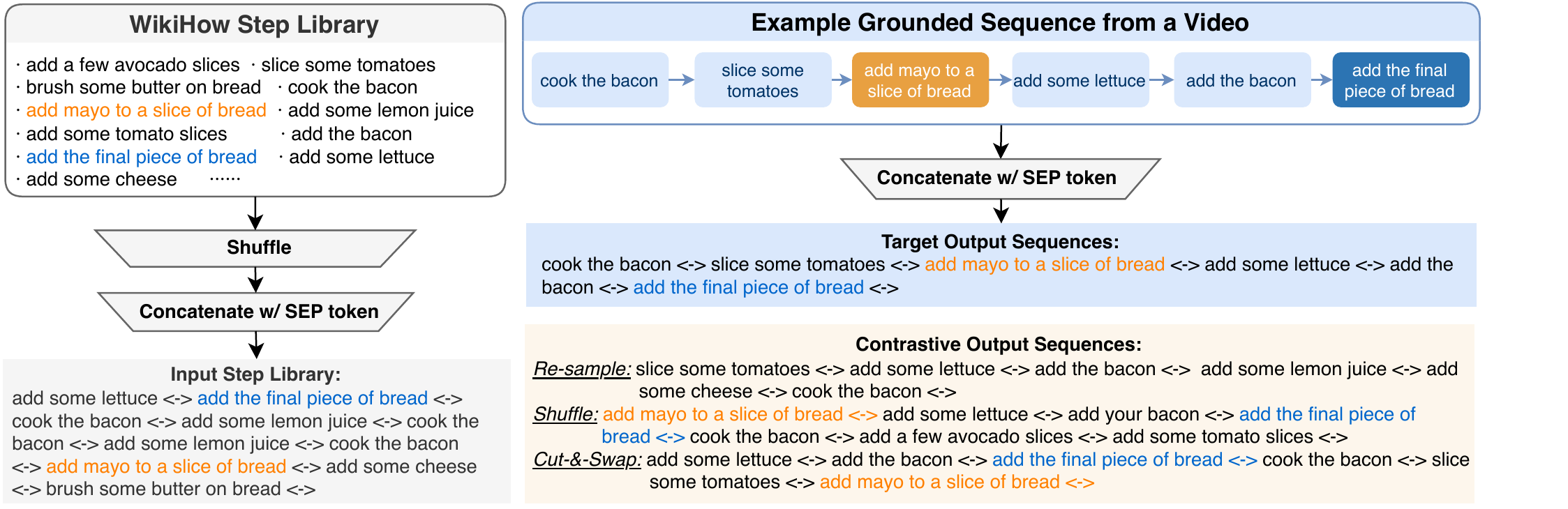}
\end{center}
\caption{
\textbf{Example input/output sequence in model training.} 
We create the input sequence by shuffling and concatenating the wikiHow step library. We use the concatenated grounded sequence as the target output (positive example) and its permuted/resampled versions as the contrastive output (negative example).
}
\label{fig:training}
\end{figure*}

\paragraph*{Processing the wikiHow Step Library:}
High quality step-level alignment demands the wikiHow Step Library used for grounding to contain clean, non-overlapping steps that are homogeneous in format and granularity to the video step annotations. Since the vanilla wikiHow dataset~\cite{DBLP:journals/corr/abs-1810-09305} does not meet these criteria, we perform a series of pre-processing before step-level alignment:

\begin{enumerate}
\itemsep0em 
    \item First, we put the steps in the initial wikiHow step library through a series of regex-based parsing to standardise stylistic elements like capitalization, punctuation and bracket/parentheses usage.
    \item Then, we use a seq2seq text simplification model~\cite{NAACL-2021-Maddela} to reduce granularity in wikiHow steps which are often more fine-grained than video step annotations. 
    \item Finally, we deduplicate the wikiHow Step Library by enforcing a minimum weighted Levenshtein distance of 0.1 between any two steps and removing overly similar duplicate steps. 
\end{enumerate}

\subsection{Model Training} \label{Script learning}
\paragraph*{Graph Script Learning} \label{section: our model}
Inspired by~\cite{DBLP:conf/icml/BojchevskiSZG18}, we transform the graph script learning problem into a path learning problem by treating steps as nodes and temporal relationships between the steps as directed edges (edges point to future step).
For each procedural task $\mathcal{T}$, the wikiHow step library of task-relevant steps $\mathcal{W_\mathcal{T}}$ generated in \S\ref{grounding_section} represents the set of nodes used to construct the latent ground-truth graph script.
In \S\ref{grounding_section}, we grounded each procedural video to a wikiHow step sequence. These step sequences can be regarded as observed step node paths that lead to successful completion of  $\mathcal{T}$. 
In this formulation, learning the latent graph script for a task can be regarded as learning the weights of valid paths through $\mathcal{W_\mathcal{T}}$.


For our basic architecture, we train a BART-base model~\cite{Lewis2019BARTDS} to generate complete step sequences given a wikiHow step library. As illustrated in Fig.\ref{fig:training}, for each task $\mathcal{T}$, we first shuffle the corresponding wikiHow step library to remove any pre-existing step ordering. Then, we concatenate the shuffled step library with a special separator token\footnote{We define the separator token as <->.} appended to the end of every step to indicate step boundary. The resulting sequence is used as the input sequence for all training data regarding $\mathcal{T}$.
For each target output, we first collect all grounded step sequences of videos completing $\mathcal{T}$. Similar to input sequences, steps in the output are also appended with the same separator token and concatenated. Finally, each processed video-grounded step sequence is used individually as a target output sequence for our model.




\paragraph*{Path-Level Constraint}  \label{contrastive_loss_section}
Besides being able to generate valid step sequences that lead to successful task completion, we also enable our model to differentiate valid step sequences from invalid ones that fail to complete the task. We accomplish this by introducing a Path-Level Constraint in the form of a contrastive loss. For each positive step sequence, we generate $n$ negative contrastive sequences using the following 3 methods (Fig.\ref{fig:training}):


\begin{enumerate}
\itemsep0em 
    \item \textit{\underline{Re-sample}:} randomly re-sample a step sequence of the same length from the wikiHow step library. Both step selection and step ordering are wrong.
    \item \textit{\underline{Shuffle}:} shuffle the sequence until no longer valid. Step selection is preserved, but local/global step ordering are wrong.
    \item \textit{\underline{Cut \& Swap}:} cut 
    the sequence at a random position and swap the latter part to the front. Step selection and local step ordering are preserved, but global step ordering is wrong.
\end{enumerate}
To maximize the model's learning potential, we follow the paradigm of curriculum learning~\cite{10.1145/1553374.1553380} when introducing contrastive examples: we start with contrastive sequences generated via \textit{\underline{Re-sample}} because they are most dissimilar from valid sequences. As training progresses, we shift toward \textit{\underline{Shuffled}} and \textit{\underline{Cut \& Swap}} by gradually increasing the probability of sampling from those contrastive sequence groups.

\begin{figure*}[ht]
\begin{center}
\includegraphics[width=160mm,scale=1]{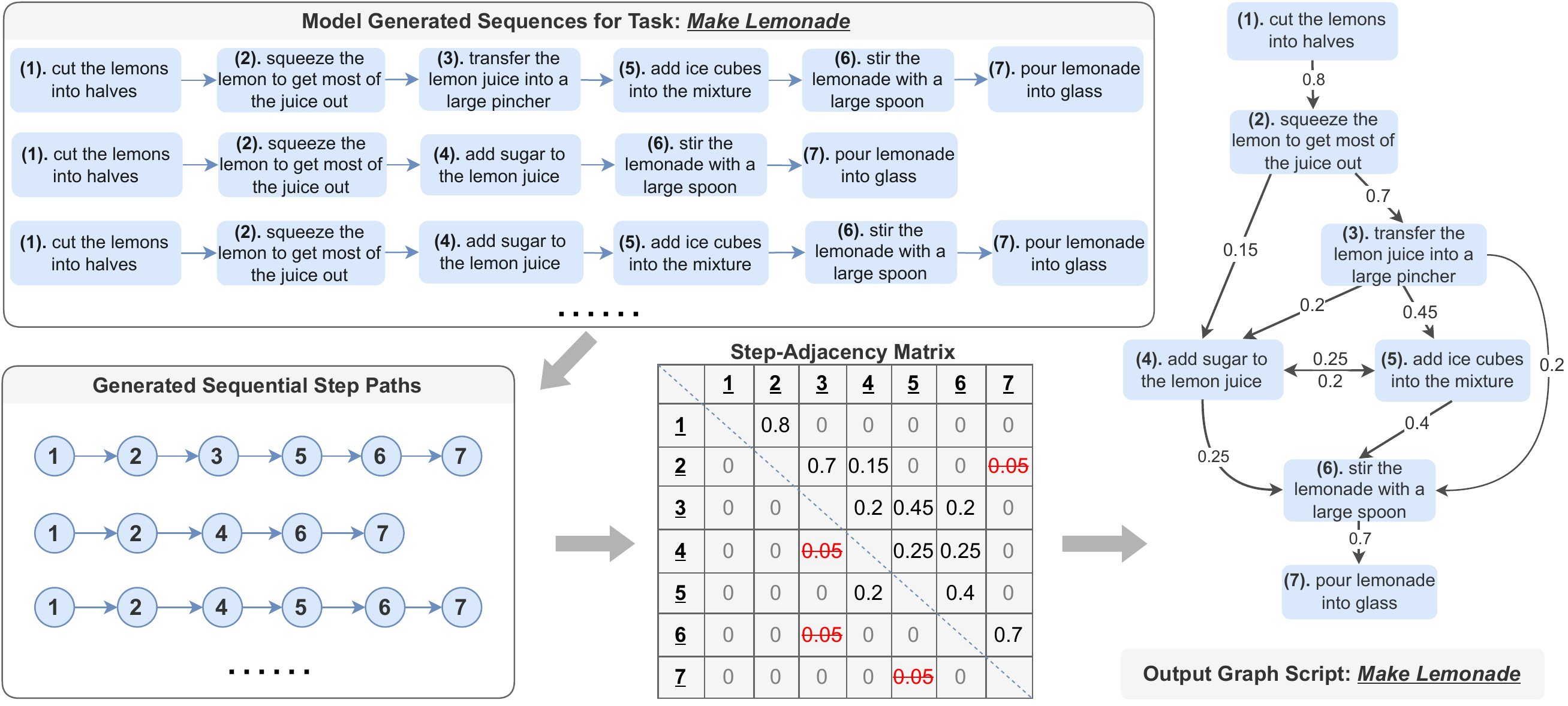}
\end{center}
\caption{\textbf{Example of Graph Script Generation.}
To decode a graph from our generator, we first ask the model to generate alternative step sequences via beam-decoding and record them in an step-adjacency matrix, which is then be used to reconstruct the non-sequential graph script (with low-frequency edges removed).
} 
\label{fig:generate graph}
\end{figure*}

Inspired by~\cite{saha2022explagraphsgen} 
, we use the last layer of the decoder in BART as the representation of each token in the sequence and obtain the sequence representation by averaging over the constituent token representations. 
Let the hidden representations of our generated sequence $\mathbf{s}^{(g)}$, true grounded sequence $\mathbf{s}^{(p)}$ and negative contrastive sequence $\mathbf{s}^{(n)} $ be denoted by $\mathbf{z}^{(g)}$, $\mathbf{z}^{(p)}$ and $\mathbf{z}^{(n)} $, respectively. Let $\mathcal{Z}=\left\{\mathbf{z}^{(p)}\right\} \bigcup\left\{\mathbf{z}_i^{(n)}\right\}_{i=1}^M$, with $M$ as the number of negative contrastive sequences. Hence, we define our Path-level Contrastive Loss: \footnote{Based on the InfoNCE Contrastive Loss\cite{DBLP:journals/corr/abs-1807-03748}}:
\begin{equation}
\resizebox{.891\hsize}{!}
{$
\mathcal{L}_{PC}=-\log \frac{\exp \left[\operatorname{sim}\left(\mathbf{z}^{\mathbf{(g)}}, \mathbf{z}^{(p)}\right) / \tau\right]}{\sum_{\mathbf{z}^{(i)} \in \mathcal{Z}} \exp \left[\operatorname{sim}\left(\mathbf{z}^{\mathbf{(g)}}, \mathbf{z}^{(i)}\right) / \tau\right]}
$}
,
\end{equation}


 where the temperature $\tau$ is a hyperparameter and $\operatorname{sim}$ denotes cosine similarity. Finally, our overall loss combines the Path-level Contrastive Loss with the Cross-Entropy Loss of seq2seq models:

\begin{equation}
\mathcal{L}_{CE}=\sum_{i}-\log P\left(\mathbf{s}_i^{(p)} \mid \mathbf{s}_{<i}^{(p)}, \mathcal{W_\mathcal{T}} \right),
\end{equation}

\begin{equation}
\mathcal{L}_{\text {total}}= \mathcal{L}_{\text {CE}} + \alpha \mathcal{L}_{\text {PC}},
\end{equation}

where $\alpha$ is a hyperparameter and $\mathcal{W_\mathcal{T}}$ denotes the task-specific wikiHow step library.



\subsection{Graph Script Generation}\label{sec:graph_generation}

In \S\ref{Script learning}, we transformed the graph script learning problem into a path learning problem by treating procedural step relationships as edges between nodes and abstracting the latent ground truth graph as the collection of paths through node-set $\mathcal{W_\mathcal{T}}$ that lead to successful task completion. After our model has learnt the latent ground truth graph scripts for a set of tasks, we use it to reconstruct explicit graph scripts through the following procedure:

For each task $\mathcal{T}$, we use $\mathcal{W_\mathcal{T}}$ as model input and have the model generate output step sequences consisting only of steps within $\mathcal{W_\mathcal{T}}$. 
We enforce this by implementing Step-constrained Beam Search, an extension of Constrained Beam Search~\cite{decao2021autoregressive}, where the model is only allowed to generate valid next words that lead to entities stemming from a fixed prefix trie $\mathcal{P}$. Here, we construct $\mathcal{P_\mathcal{T}}$ containing all steps in $\mathcal{W_\mathcal{T}}$ and ask the model to repeatedly decode from $\mathcal{P_\mathcal{T}}$ to generate step sequences. After each step is fully generated, the model is given the choice to end generation by producing the end-of-sentence (eos) token or continue decoding the next step by producing a token from the root of $\mathcal{P_\mathcal{T}}$. 
After generating the predicted step sequences, we break them down and record the edges in an graph adjacency matrix between all generated step nodes. The low-frequency edges representing unlikely paths are removed to improve graph script confidence.  Finally, we reconstruct the output graph script from the graph adjacency matrix. An example of this process on the task \underline{\textit{"Make Lemonade"}} is detailed in Fig.\ref{fig:generate graph}.

\begin{table*}
\setlength\tabcolsep{7.3pt}
\setlength\extrarowheight{1.2pt}
\small
\centering
\begin{tabular}{l c|c c c c|c c c}
\toprule
\multirow{3}{*}{\textbf{Model}} & \multirow{3}{*}{HT100M}& \multicolumn{4}{c|}{\textbf{Next Step Prediction}} & \multicolumn{3}{c}{\textbf{Partial Sequence Completion}} \\
& & Acc@1 & Acc@3 & Rec@3 & F\textsubscript{1}@3 & Acc@1 & Edit & Normalized \\
& &$\uparrow$ & $\uparrow$ & $\uparrow$ & $\uparrow$ & $\uparrow$ & Dist. $\downarrow$ & Edit Dist. $\downarrow$\\
\midrule
TimeSformer+DS & \xmark & 59.91 & 60.82 & 52.98 & 43.83 & - & - & - \\
\midrule
Random & \xmark & 31.34 & 50.32 & 28.84 & 38.04 & 1.20 & 2.398 & .6935 \\
wikiHow Linear & \xmark & 44.05 & 59.51 & 54.02 & 42.14 & 11.74  & 1.872 & .6061 \\
ReBART & \xmark & 49.07 & 58.00 & 61.39 & 44.38 & 18.28 & 1.802 & .4411 \\
Direct NSP (Grounding) & \xmark & 68.89  & 63.02 & 79.01 & 53.85 & - & - & -  \\

Direct PSC (Grounding) & \xmark & - & - & - & - & 29.17 & 1.214 & .4118  \\

\hline\rowcolor{gray!20}
Ours (Grounding) & \xmark & 75.59 & 67.50 & \textbf{83.17} & 58.29 & 20.12  & 1.639 & .4296  \\
\rowcolor{gray!20}
\rowcolor{gray!20}
Ours (Grounding) & \cmark & 70.97  & \textbf{74.68} & 74.14 & 61.52 & 29.34 & 1.193 & .4093 \\
\rowcolor{gray!20}
Ours (Grounding + PLC) & \xmark & 75.49 & 71.89  & 72.51 & 58.48 & 26.70 & 1.228 & .4267 \\
\rowcolor{gray!20}
Ours (Grounding + PLC) & \cmark & \textbf{76.09}   & 73.72 &  78.22 & \textbf{61.90} &\textbf{32.08}  & \textbf{1.123} & \textbf{.3849} \\
\Xhline{2\arrayrulewidth}
\end{tabular}

\caption{\textbf{
Automatic Evaluation Results on Next Step Prediction and Partial Sequence Completion.}  Here ``HT100M'' denotes whether the model is pre-trained on the Howto100M dataset with temporal order information. ``{Normalized Edit Dist.}'' represents the average Levenstein distance normalized by sequence length. ``Grounding'' denotes whether the model used our grounded video sequences for training. ``PLC'' represents Path-Level Constraint.
}
\label{tab:automatic_eval}
\end{table*}

\section{Experiments}

To evaluate our non-sequential graph script induction model, we propose 3 new downstream tasks:

\begin{enumerate}
\itemsep0em 
    \item Graph Script Generation: for each task $\mathcal{T}$, the system is asked to produce a 2-dimensional probabilistic graph script similar to Fig.\ref{fig:schema} that captures the step relationships introduced in section \ref{intro_section}). The model is scored based on human evaluation of its generated graph scripts.
    \item Next Step Prediction: given a partial step sequence $S_p = (s_1 \rightarrow ... \rightarrow s_{t-1} )$, the model is asked to predict the top-k most likely choices for the next step $s_t$ from $\mathcal{W}_\mathcal{T}$. For each partial step sequence, there can be a variable number of correct next steps.
    \item Partial Sequence Completion: given a partial step sequence $S_p = (s_1 \rightarrow ... \rightarrow s_{t-1} )$, the model is asked to produce a sequence $S = (s_1 \rightarrow ... \rightarrow s_n )$ using steps from $\mathcal{W}_\mathcal{T}$ that completes the task $\mathcal{T}$. This task is particularly challenging because the model is asked to predict a variable-length step sequence that best completes the task at hand.
\end{enumerate}



\subsection{Baselines} \label{sec:Baselines}


\noindent
\textbf{Baseline: TimeSformer+DS.} TimeSformer~\cite{gberta_2021_ICML} trained with unsupervised distant supervision~\cite{lin2022learning} provides the state-of-the-art step-level video representation for pure-video-based step forecasting. We fine-tuned the model on CrossTask videos before testing.

\noindent
\textbf{Baseline: wikiHow Linear.}
This model is trained on all wikiHow linear step sequences selected during title-matching (\S\ref{grounding_section}). 
To ensure fairness in comparison, the training sequences undergo the same step processing as that of the non-sequential model. 
For each training sequence, the model takes the complete wikiHow step library as input and one linear sequence from the selected wikiHow documents as target output.

\noindent
\textbf{Baseline: ReBART.} ReBART~\cite{DBLP:journals/corr/abs-2104-07064} is the state-of-the-art sentence re-ordering method that uses a text-to-marker generation format. Numbered markers are inserted before each step in the training data, and the target output step sequence is translated into corresponding marker sequences.

\noindent
\textbf{Ablation Study: Direct Next Step Prediction \& Direct Partial Sequence Completion.} These two task-specific models are included as variants of our model (\S\ref{section: our model}) where the input training sequence is a partial start sequence and the target output sequence is just the next step (for next step prediction) or the remaining sequence (for partial sequence completion). The training data for these two models are also constructed from our grounded video step sequences (\S\ref{grounding_section}).

\subsection{Automatic Evaluation} 

\paragraph*{Evaluation Dataset}
Inspired by ~\cite{chen2022weakly}, we build our evaluation dataset on top of the existing CrossTask Dataset~\cite{Zhukov2019CrossTaskWS} and reuse their manual temporal step annotations. 
Using procedures in \S\ref{grounding_section}, we ground annotated CrossTask videos
(Fig.\ref{fig:grounding}) to sentence-simplified wikiHow Steps.
Afterwards, we randomly select 40\% of grounded step sequences to form the training set. Remaining sequences form the test set.

For each grounded step sequence $S = (s_1 \rightarrow ... \rightarrow s_n )$ in the test set, we split after all steps $(s_t | t \in [1,n-1] )$ to produce partial start sequences $S_p = (s_1 \rightarrow ... \rightarrow s_{t} )$. For next step prediction, the correct output corresponding to $S_p$ is the next step $s_{t+1}$; while for partial sequence completion, the correct output corresponding to $S_p$ is the remaining sequence $(s_{t+1} \rightarrow ... \rightarrow s_n)$. In the case where multiple grounded step sequences share the same partial start sequence $S_p$ but have different next step / remaining steps, the input sequence $S_p$ would have multiple correct answers for next step prediction / partial sequence completion.

\paragraph*{Next Step Prediction}
As shown in Table~\ref{tab:automatic_eval}, our models trained using video-to-text grounded step sequences outperform other baselines trained with wikiHow linear step sequences by 15\% $\sim$ 20\% absolute gains in all next step prediction metrics. This shows the advantage of our video-grounded step sequences over wikiHow linear sequences in improving the model's ability to predict next steps.
Comparing our models trained on complete step sequences against models trained directly on next step prediction without whole script knowledge, we see a large performance gap.
This shows the importance of learning whole-script knowledge for next step prediction.
When predicting top-3 most likely next steps, models pretrained on Howto100M significantly outperform models w/o pretraining. This can be attributed to the pretrained models having better knowledge of sequence "branching" from observing more diverse task executions.

\paragraph*{Partial Sequence Completion}
Our best performing models trained using video-to-text grounded step sequences typically achieves over 13\% absolute gains on Accuracy@1 and over 14\% relative gains on normalized edit distance against other baselines trained using wikiHow linear step sequences, showing grounded videos step sequences can boost models' ability in partial sequence completion.
When comparing models trained with the Path-Level Constraint (Sec.\ref{contrastive_loss_section}) to otherwise identical models trained without such constraint, we see significant gains across all metrics. This demonstrates the effectiveness of our Path-Level Constraint in teaching the model to produce valid step sequences while avoiding their invalid counterparts.
We also observe a performance gain for models pretrained on Howto100M vs the same models w/o such pretraining. This result combined with similar results in next step prediction shows that pretraining on a large unlabelled procedural video dataset can improve the model's ability to learn scripts for other tasks.

\begin{table}
\setlength\tabcolsep{7.5pt}
\setlength\extrarowheight{1.2pt}
\small
\centering
\begin{tabular}{l | c c | c c }
\toprule
\multirow{2}{*}{\textbf{Relation Type}} & \multicolumn{2}{c|}{\textbf{Linear}} & \multicolumn{2}{c}{\textbf{Ours}} \\
 & \#/task & Acc & \#/task & Acc  \\

\midrule
Sequential & 10.56 & 35.79 & 12.50 & 88.02 \\
Optional & 1.40 & 19.23 & 2.44 & 65.91 \\
Interchangeable & 0.44 & 37.50 & 1.44 & 88.46 \\
\midrule
Overall & 12.40 & 33.93 & 16.38 & 82.69 \\

\Xhline{2\arrayrulewidth}
\end{tabular}

\caption{Human Evaluation results by step-relation type.}
\label{tab:human_eval_relation}

\end{table}

\begin{table}
\setlength\tabcolsep{7.5pt}
\setlength\extrarowheight{1.2pt}
\small
\centering
\begin{tabular}{l | c c | c c }
\toprule
\multirow{2}{*}{\textbf{Task Category}} & \multicolumn{2}{c|}{\textbf{Linear}} & \multicolumn{2}{c}{\textbf{Ours}} \\
 & \#/task & Acc & \#/task & Acc  \\

\midrule
Cooking & 12.1 & 35.16 & 16.2 & 81.07 \\
Household & 12.5 & 28.33 & 16.0 & 75.00 \\
Car Maintenance & 15.0 & 36.67 & 17.5 & 88.89 \\

\Xhline{2\arrayrulewidth}
\end{tabular}

\caption{Human Evaluation results by task category. \#/task denotes average number of relations per task.}
\label{tab:human_eval_category}

\end{table}

\subsection{Human Evaluation}\label{sec:human_eval}

Using the graph construction method in \S\ref{sec:graph_generation}, we generate two graph scripts for each procedural task in CrossTask using the wikiHow Linear baseline (\S\ref{sec:Baselines}) and our non-sequential graph script induction model. To evaluate the correctness and expressiveness of generated graph scripts, we design T/F questions regarding \textit{sequential}, \textit{optional}, and \textit{interchangeable} relations. For optional and interchangeable step relationships indicated by the graph script, we ask annotators whether the relationship is appropriate. For other steps in the connected graph script, we ask annotators whether their previous and subsequent steps are sequentially appropriate.

Table~\ref{tab:human_eval_relation} and table~\ref{tab:human_eval_category} show our model achieves 46.68\% $\sim$ 52.23\% absolute gains in Accuracy across all relation types and task categories. In addition, our model is able to accurately capture 74\% more optional steps and 227\% more interchangeable step pairs in generated graph scripts.

\section{Related Work}

\paragraph*{Text-based Script Induction}
Temporal relations have always been the core of script (schema) related tasks, which can either be learned from data or human annotation. 
When human-written scripts are available, previous works have typically assumed that the human-provided ordering of steps is the only correct order~\cite{Jung2010AutomaticCO, Ostermann2017AligningSE,Nguyen2017SequenceTS, LYU2021GoalOrientedSC, sakaguchi-etal-2021-proscript-partially}. 
Another line of work has attempted to learn event ordering from data alone, either by assuming that the events follow narrative order~\cite{chambers-jurafsky-2008-unsupervised, Chambers2009UnsupervisedLO, Jans2012SkipNA, Rudinger2015ScriptIA, Ahrendt2016ImprovingEP,Wang2017IntegratingOI} or by using an event-event temporal relation classifier to predict the true ordering of events~\cite{Li2020ConnectingTD,Li2021TheFI}.
Our work is distinct from both paradigms as we use human-written scripts as a basis and learn the event ordering from observed sequences in videos.

\paragraph*{Video-based Script Induction}
Existing efforts that utilize visual information in script induction can be mainly classified into implicit script knowledge models and explicit sequential script induction models. Some previous efforts have focused on training models with implicit script knowledge that can make step-level predictions based on textual~\cite{Yang2021VisualGI}, visual~\cite{Sener2018ZeroShotAF, lin2022learning, Zhou2023ProcedureAwarePF}, or multimedia~\cite{Zellers2021MERLOTMN} input. Other models aim to produce explicit sequential graph scripts that only capture procedural relations between steps~\cite{Salvador2018InverseCR, Yang2021InduceER}. Another line of works use multimedia information to generate explicit graph scripts that model only pre-conditional/dependency relationships between events~\cite{Logeswaran2023UnsupervisedTG} and sub-events~\cite{Jang2023MultimodalSG}. Ours is the first work to generate explicit non-sequential graph scripts that capture rich procedural, optional, and interchangeable relations through multimedia learning.





\section{Conclusions and Future Work}

We propose the new task of Non-sequential Graph Script Induction to capture optional and interchangeable steps in procedural tasks. Instead of relying on the script annotation, we automatically induce graph scripts by grounding procedural videos to a wikiHow textual step library. We transform the graph generation problem to a path generation problem that can better aligned with video observations, and train a seq2seq model using our grounded step sequences while imposing path-level constraints via a contrastive loss. Experiments demonstrate our model's superiority on downstream tasks including next step prediction and partial sequence completion. Human evaluation confirms our model's ability to generate  graph scripts that correctly capture optional and interchangeable steps. 
Future work will focus on incorporating more video supervision signals such as enriching steps from videos and adding the repeatable steps. 

\section{Limitations}

\subsection{Representation of Repeatable Steps}

Our current approach is not able to capture repeatable steps due to data source constraints from our video datasets. The video datasets we use in this work, namely Howto100M and CrossTask, are both constructed from Youtube videos. 
At the end of many Youtube instructional videos, there is a brief recap of the whole task, where many steps are displayed for a second time. 
Since CrossTask was originally proposed for step segmentation, the step annotations capture all video references to task-related steps, including the brief mentions at the end of the videos that are not actually part of task execution. Similarly, Howto100M videos ASR pieces near the end of the video would also capture the vioceover going through such step references.

Therefore, to ensure the grounded video step sequence only contains steps included in the execution of the task, we simply removed all repeated steps in the grounded step sequence and only kept the first occurrence.
However in this process, we also removed valid repeats of the same step. For example, if the step \texttt{Add some salt} was executed twice at different stages of the task. We leave this area of improvement for future works.


\subsection{Enrichment of Steps from Video}

In our current model, all the steps in the wikiHow step library are processed steps from related wikiHow documents. However, it has been shown that textual sources can be prone to reporting bias, occasionally ignore task-relevant information that is present only in the vision modality~\cite{Chen2021JointME}. 

Continuous frames from video data can capture details that text descriptions do not explicitly mention. If the model is able to make use of such vision-exclusive details and learn patterns from them, its overall ability can be improved. The challenge in utilizing such underlying visual information is to differentiate task-relevant video steps from their task-irrelevant counterparts. This area has not been covered by our current graph script induction pipeline, we hope to provide comprehensive solutions in future work.


\section{Ethics and Broader Impact}

\subsection{Datasets}
In this work, we used publicly available text data from the wikiHow Dataset (\url{https://github.com/mahnazkoupaee/wikiHow-Dataset}) Creative Commons License (CC-BY-NC-SA), which is under the \textit{Attribution-Noncommercial-Share Alike 3.0 Creative Commons License} which allows us to use the dataset for non-commercial purposes.
For video data, we used the publicly available CrossTask Dataset (\url{https://github.com/DmZhukov/CrossTask}) under the BSD 3-Clause "New" or "Revised" License and the Howto100M Dataset (\url{https://www.di.ens.fr/willow/research/howto100m/}) under the Apache License 2.0. Both licenses allows us to use the datasets for non-commercial purposes.

The datasets we use consist of non-offensive instructional and procedural videos / text scripts about everyday tasks. Our usage of the datasets only concerns the task related information and does not violate privacy.

\subsection{Human Evaluation}
As detailed in \S\ref{sec:human_eval}, we conduct human evaluation for our generated graph scripts in this paper via Amazon Mechanical Turk(\url{https://www.mturk.com/}). All annotators involved in the human evaluation are voluntary participants and receive a fair wage. All annotators were instructed of the task nature and consent to complete the annotation via a online consent form. We have applied for IRB exemption and the request was approved.

\subsection{Model Usage}
Our graph script induction framework is not intended to be used for any activity related to any human subjects. Instead, it should only be used for generating graph scripts regarding everyday tasks that benefit people's learning and understanding. It may also be used for predicting/instructing future step/steps to facilitate completion of relevant tasks. Note that our graph script induction framework is intended for wikiHow visual tasks and might not be applicable for other scenarios.


\section*{Acknowledgement}
We thank the anonymous reviewers for their helpful suggestions. This research is based upon work supported by U.S. DARPA KAIROS Program No. FA8750-19-2-1004. The views and conclusions contained herein are those of the authors and should not be interpreted as necessarily representing the official policies, either expressed or implied, of DARPA, or the U.S. Government. The U.S. Government is authorized to reproduce and distribute reprints for governmental purposes notwithstanding any copyright annotation therein.

\bibliography{ref}
\bibliographystyle{acl_natbib}

\clearpage
\appendix
\section{Appendix}

\subsection{Grounding Details}

 The following hyper-parameters used in the grounding section are determined empirically. In video to text grounding, for each video, we find the top-10 most relevant wikiHow documents. For keyword matching at the task level, we first select wikiHow documents whose title contains $\ge$ 85\% of keywords in the task name. This is to avoid calculating Sentence-BERT similarity between the task name and all wikiHow document titles. If this does not yield $\geq$ 10 documents, we relax the threshold to 75\%. For title matching and step matching, the Sentence-BERT similarity thresholds are determined empirically by qualitative evaluation over a small subset of 150 examples. For labelled videos, the step-level grounding similarity threshold $k_1$ is 0.35. For unlabelled videos, the task-level grounding similarity threshold $k_2$ is 0.75 and the step-level grounding similarity threshold $k_3$ is 0.40.

\subsection{Training Details}

For our models and baselines, we mainly use the BART-base model (140M Parameters) from the Huggingface Framework~\cite{Wolf2019HuggingFacesTS}\footnote{\url{https://huggingface.co/docs/transformers/index}}. We normalize all input and target output sentences into lower case and remove special non-English characters. For training, we use the AdamW optimizer~\cite{Loshchilov2017DecoupledWD} with a learning rate of $2\times 10 ^{-5}$ and 1000 warm-up steps. We use max input and output sequence length of 1024 for training and testing. 

For the InfoNCE contrastive loss, we set the temperature $\tau=0.1$. To implement curriculum learning. We reset the probability of sampling from different contrastive sequence groups every 5 epochs. At first we only use 're-sampled' contrastive sequence, then in every 5 epochs we transfer 20\% probability to sampling from the 'shuffled' contrastive sequences. After 25 epochs, we starting the same shift from 'shuffled' contrastive sequences to 'cut \& swapped' contrastive sequences.

We use NVIDIA V-100 GPUs with 16GB RAM and full precision. Due to GPU RAM limitation, we use gradient accumulation with equivalent batch size of 32. Training our basic model on the CrossTask training set takes approximately 5 hours while training our contrastive model with Path-Level Constraint will take 20 hours. Pre-training our model on Howto100M grounded sequence takes approximately 3 days.

\subsection{Inference Details}
During graph generation, for Step-constrained Beam Search, we use a beam number of 40 to sample steps sequences for producing the graph script. Afterwards, we filter out low-frequency edges in the adjacency graph with edge weight $\leq 0.175$ (or in this case occurrence $\leq 7$). The remaining edges are used to construct the final graph script.

\subsection{Human Evaluation Details} \label{sec:human_eval_detail}

In our human evaluation of model-generated graph scripts, three types of questions are asked regarding corresponding types of step relationships as displayed in the generated graph script:

\begin{enumerate}
    \item \textit{\underline{Optional}:} Do you think step (a) is optional when completing this task?
    \item \textit{\underline{Interchangeable}:} Do you think the steps (b) and (c) are interchangeable (can be executed in either order) when completing this task?
    \item \textit{\underline{Sequential}:} Do you think the previous and/or subsequent steps for step (d) are reasonable when completing this task?
\end{enumerate}

\begin{figure*}[!ht]
\begin{center}
\includegraphics[width=160mm,scale=1]{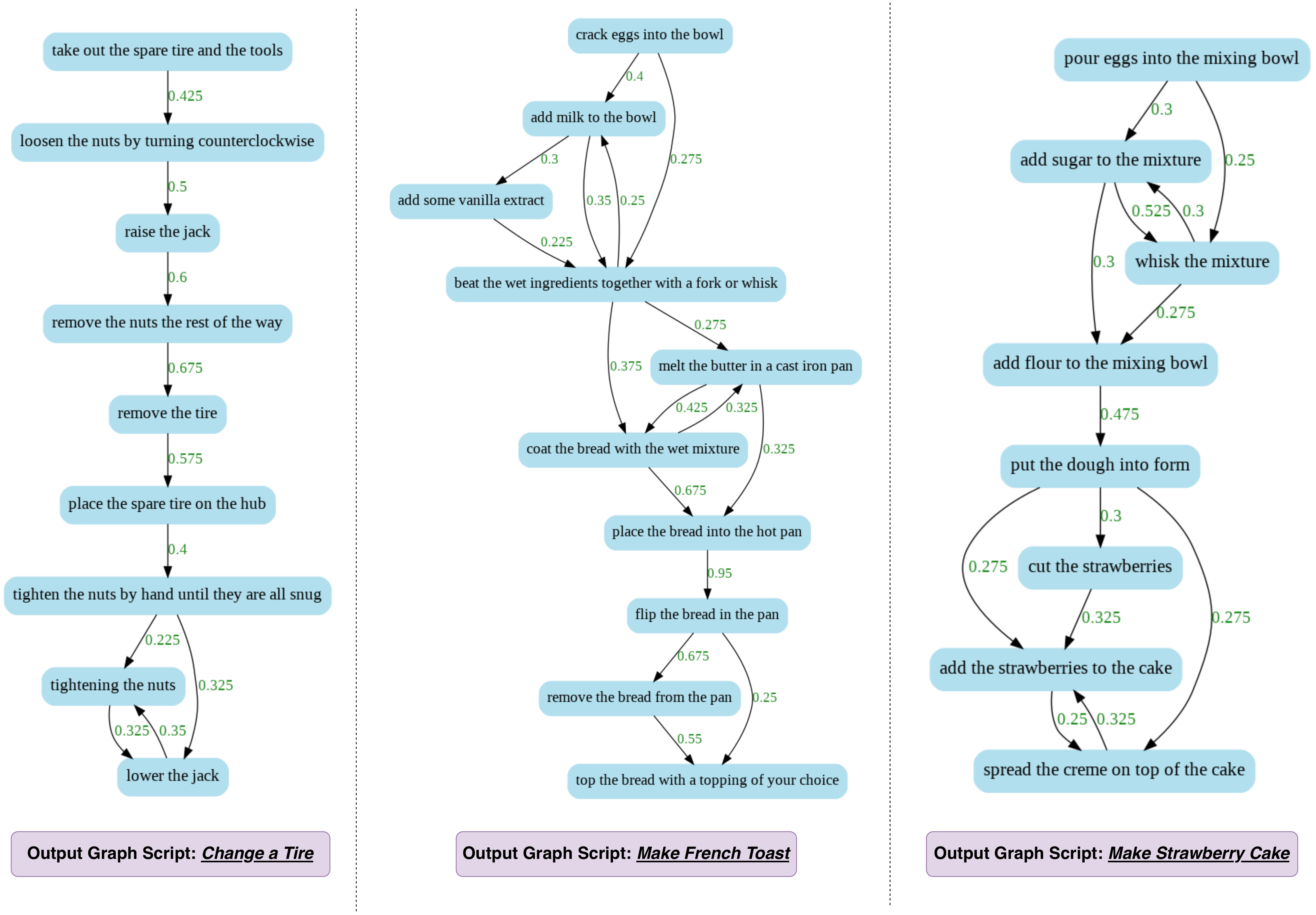}
\end{center}
\caption{Example outputs of our non-sequential graph script induction model used in Human Evaluation}
\label{fig:human_long}
\end{figure*}

To make questions more direct and objective for the annotators, each question only focuses on a small portion of steps in the generated graph script. For example, given the output graph script for the task "Make Strawberry Cake" as shown in Fig.\ref{fig:human_long}, the annotator would be asked the following questions (partial):

\begin{enumerate}
    \item \textit{\underline{Optional}:} Make Strawberry Cake: Do you think the step "cut the strawberries" is optional when completing this task?
    \item \textit{\underline{Interchangeable}:} Make Strawberry Cake: Do you think the steps "add sugar to the mixture" and "whisk the mixture" are interchangeable (can be executed in either order) when completing this task?
    \item \textit{\underline{Sequential}:} Make Strawberry Cake: Do you think the previous and/or subsequent steps for step "add flour to the mixing bowl" are reasonable when completing this task?
\end{enumerate}

\begin{figure*}
    \centering
    \includegraphics[width=\linewidth]{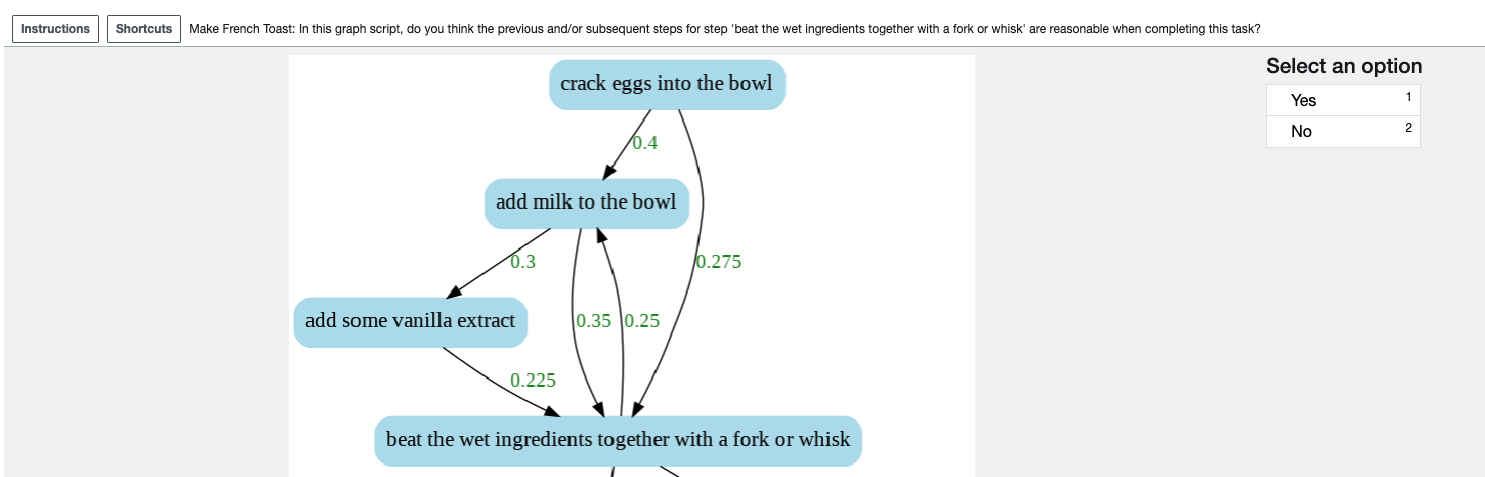}
    \caption{A screenshot from our annotation task on Mechanical Turk.}
    \label{fig:mturk_screenshot}
\end{figure*}

We used 8 human annotators while each annotator answered (on average) 65 questions. Each question is assigned to $\geq$ 2 annotators with 72.31\% inter-annotator agreement. An example screenshot of the annotation interface is shown in Fig.\ref{fig:mturk_screenshot}.

\end{document}